\pdfoutput=1

\documentclass[opre,nonblindrev]{informs3}

\OneAndAHalfSpacedXII

\usepackage{natbib}
 \bibpunct[, ]{(}{)}{,}{a}{}{,}%

 \usepackage{wrapfig,lipsum,booktabs, makecell}
\newcommand{\sd}[1]{{\scriptsize\color[gray]{0.45}$#1$}}
\makeatletter\def\theARTICLETOP{\vspace*{-36pt}}\makeatother

\TheoremsNumberedThrough     
\ECRepeatTheorems

\EquationsNumberedThrough    

\MANUSCRIPTNO{}

\usepackage{psfrag,ifthen,multirow,mathrsfs}
\usepackage{epsfig,psfrag,subcaption,float,texnansi,color,tikz,ifthen}
\usepackage{algorithm}
\usepackage{algpseudocode}
\usepackage{microtype}
\usepackage{graphicx}
\usepackage{latexsym,url,afterpage,enumerate,lmodern}
\usepackage[normalem]{ulem}
\usepackage{amsmath,amssymb,amsfonts}
\usepackage[margin=10pt,font=small,labelfont=bf]{caption}
\usepackage{booktabs,xcolor}
\usepackage{IEEEtrantools}
\usepackage{bm}
\usepackage{enumitem}
\usepackage{tabularx}
\usepackage{dsfont}
\usepackage{rotating}
\usepackage{lscape}
\usepackage{wrapfig}
\usepackage{mathtools}
\usepackage[table]{xcolor}

\newcommand{\revision}[1]{{\normalfont #1}}

\usepackage{changepage}

\makeatletter
\renewcommand{\theARTICLEABSTRACT}{%
  \begin{Center}
  \HOOKb
  \vspace*{18pt}
  \end{Center}
  \begin{adjustwidth}{1.5pc}{1.5pc}\parindent1em
    \ABSfont
    \noindent\theABSTRACT\endgraf
    \vskip5pt
    \theKEYWORDS
    \theSUBJECTCLASS
    \theAREAOFREVIEW
    \theMSCCLASS
    \theORMSCLASS
    \if@BLINDREV\else
	\fi
  \end{adjustwidth}}
\makeatother

\begin{document}


\RUNAUTHOR{Edmonds et al.}

\RUNTITLE{Learning Risk Scores Robust to Unobserved Confounders}

\TITLE{Learning Risk Scores Robust to Unobserved Confounders}

\ARTICLEAUTHORS{%
\AUTHOR{Ryan Edmonds\textsuperscript{a,b,}*, Yingxiao Ye\textsuperscript{a,c}, Sina Aghaei\textsuperscript{a,c}, Andr\'es G\'omez\textsuperscript{a,c}, \c{C}a\u{g}{\i}l Ko\c{c}yi\u{g}it\textsuperscript{d}, \\ Phebe Vayanos\textsuperscript{a,b,c}}
\AFF{\textsuperscript{a}Center for Artificial Intelligence in Society, University of Southern California, Los Angeles, CA 90089, USA\\
\textsuperscript{b}Department of Computer Science, University of Southern California, Los Angeles, CA 90089, USA\\
\textsuperscript{c}Department of Industrial and Systems Engineering, University of Southern California, Los Angeles, CA 90089, USA\\
\textsuperscript{d}Department of Engineering, Department of Economics and Management, University of Luxembourg, L-1359 Luxembourg\\[2pt]
*Corresponding author: \EMAIL{ryanedmo@usc.edu}}
}

\ABSTRACT{
We consider the problem of learning risk scores to prioritize individuals for scarce resources or interventions, from historical observational data affected by unobserved confounding. In settings such as public health and homelessness prevention, decisions about who receives a scarce resource (e.g., a hospital bed or housing) are often guided by a risk score assigned to each individual based on recorded characteristics, such as responses to a survey. These risk scores are increasingly being learned directly from observational data: historical records of individuals' characteristics, allocation decisions, and outcomes under received allocations. Standard methods such as inverse propensity weighting (IPW), which corrects for the bias introduced by the historical allocation policy, can be used to learn accurate risk scores if the historical decision process is fully explained by the recorded characteristics, an assumption known as unconfoundedness. In practice, however, historical decisions often depend on unrecorded information (e.g., details a caseworker learns in conversation but never logs), causing learned risk scores to systematically under-prioritize exactly the individuals whose unrecorded circumstances drove past prioritization. We propose a method for learning risk scores that are robust to this kind of unobserved confounding, building on IPW. Since propensity weights cannot be reliably estimated under unobserved confounding, we instead treat them as belonging to an uncertainty set determined by the observable data and domain-informed estimates of the degree of confounding, combining sensitivity analysis from causal inference with Wasserstein distributionally robust optimization. The resulting robust risk score learning problem admits a sample-based approximation that we reformulate as an exponential cone program compatible with off-the-shelf solvers. We demonstrate the effectiveness of our approach relative to IPW and other benchmarks on semi-synthetic data derived from datasets in the UCI Machine Learning Repository. 
Our method decreases calibration error by up to $28.9\%$ over traditional benchmarks and up to $9.8\%$ over the state of the art, without compromising other metrics.
}
\maketitle
\section{Introduction}

\revision{In high-stakes resource allocation settings, decisions about who receives scarce resources are typically made by people — caseworkers, judges, clinicians — often guided by hand-crafted risk scores: mappings from individual characteristics to a measure of vulnerability that allow interpretable comparisons between different individuals' needs. Risk scores of this kind have been deployed in fields such as criminal justice and public health, where they inform consequential decisions about parole and medical treatment \citep{angwin2016compas, nemati2018sepsis}. Risk scores have traditionally been hand-crafted from domain expertise. In homelessness prevention, for example, housing authorities use standardized vulnerability surveys such as the VI-SPDAT \citep{orgcode2015vispdat}, which score risk using hand-crafted rules for different sections of questions (e.g., if a participant answers YES to any of questions 1 through 4, increase risk by 1). In recent years, however, there is a growing interest in using machine learning to learn risk scores directly from historical data, with the goal of producing scores that are more accurate than what hand-crafted rules alone can achieve \citep{Rice2023CESTTRR, Wang2023Recidivism}.}

\revision{These approaches often rely on observational data — historical records of individuals' characteristics, the allocation decisions that were actually made under some prior policy, and the outcomes that followed — rather than data from randomized assignment. In the case of homelessness, this means historical records of individuals who filled out the VI-SPDAT, whether or not they received housing, and their observed outcomes in the system (e.g., remaining housed vs. returning to homelessness). Crucially, we only observe outcomes for each individual for the specific resource they were allocated, and do not observe other counterfactual outcomes. To learn accurate risk scores from observational data, existing methods typically work by debiasing the historical allocation policy itself. A standard tool for this is inverse propensity weighting \citep{rosenbaum1983propensity}, which reweighs observations by an estimated propensity score, representing the probability an individual would have received the resource under the historical policy given their recorded characteristics. This causes the reweighted data to behave as if historical allocation had been random.}

\revision{These methods, however, typically require an unconfoundedness assumption that, conditional on the recorded characteristics, the historical allocation decision is independent of the outcome. In other words, everything relevant to both the decision and the outcome should be captured in the data. In practice, this assumption frequently fails, because historical decisions were often based on more information than was recorded. In homelessness prevention, for instance, the allocations we observe were shaped not only by survey responses but by factors caseworkers observed in person; factors such as social connections or perceived "resilience" may have no direct analogue in the recorded data. This means individuals who received housing may be systematically different from those who did not, in ways we cannot observe ex post — so a risk score learned using a method that assumes unconfoundedness may not properly debias the historical allocation policy. This problem is not unique to homelessness: whenever historical policies depended on richer information than what was recorded, propensity scores like the one above cannot be reliably estimated, because they are themselves confounded by unobserved variables.}

This gives rise to our research question: \textit{how can we learn risk scores that are robust to unobserved confounding?} 
We treat the unidentifiable propensity scores as uncertain and maximize worst-case expected log-likelihood over an uncertainty set that (i) bounds confounding via the marginal sensitivity model of \citet{tan2006msm}, (ii) enforces consistency with the observable data via a Wasserstein constraint, and (iii) can capture domain-specific logic such as historical rationality. The sample-based problem can be reformulated as an exponential cone program, and we evaluate it on semi-synthetic data from the UCI ML repository.

\subsection{Related Work}

\textbf{Risk score learning as optimization.} Risk scores are simple, additive scoring rules that let practitioners estimate risk with a handful of arithmetic operations, and have long been used in medicine and criminal justice \citep{angwin2016compas, nemati2018sepsis}. \citet{ustun2019learning} argue that such scores, which were traditionally built ad hoc through expert judgment, are better obtained by directly formulating risk score construction as an optimization program that jointly balances calibration, sparsity, and operational constraints rather than treating them as separate heuristic stages. Our work adopts this same premise — that risk scores are best learned by directly optimizing an objective that reflects the desired properties of the score — but \revision{we aim to learn risk scores from observational data, so we only observe outcomes for individuals under the treatment they received. Since treatment assignments may be subject to unobserved confounding,} we therefore extend this optimization-based framing to a robust formulation, optimizing against a worst-case realization of the confounder within a causal uncertainty set, rather than against a single fixed dataset. \revision{While our experiments focus on real-valued risk scores, our framework can easily accommodate integrality or sparsity constraints if desired.}

\textbf{Sensitivity analysis, causal uncertainty sets, and partial identification.} 
\revision{Sensitivity analysis for unmeasured confounding originates with \citet{cornfield_smoking_1959} and \citet{rosenbaum2002observational}, and was recast by \citet{tan2006msm} as the marginal sensitivity model (MSM), which bounds the difference in odds ratios between the true and observed propensity scores. \citet{bertsimas2023distributionallyrobustcausalinference} connect the MSM to distributionally robust optimization, deriving sharp average treatment effect bounds by optimizing over the induced uncertainty set; \citet{DornGuoQuantileBalancing} sharpen this characterization by reducing average treatment effect bounds to a single quantile regression, and \citet{Dorn2025DoublyValid} establish closed-form solutions for expectation-type objectives under this set. \citet{zhang2025enhancedmarginalsensitivitymodel} propose the enhanced MSM (eMSM), supplementing the MSM treatment sensitivity constraint with an outcome sensitivity constraint that bounds how much an unobserved confounder can shift treatment effects, yielding tighter sharp bounds on causal estimands. Our uncertainty set also builds on the MSM, but is shaped by a different modeling goal: rather than bounding a causal estimand such as the average treatment effect, we optimize a predictive risk score. Our formulation can accommodate general outcome sensitivity constraints in the style of eMSM, bounding the difference between true and observed treatment effects by an arbitrary interval. Historical rationality — the assumption that decision-makers, given access to richer information than what is recorded, used it on average to make better decisions than they would have without it — corresponds to the one-sided choice of this interval $[0, \infty)$. We adopt this realization specifically because, unlike eMSM's outcome constraint, it requires no additional sensitivity parameter to elicit or estimate.}

More broadly, a large literature derives bounds on causal estimands under a variety of sensitivity models \citep{tan2006msm, DornGuoQuantileBalancing, Dorn2025DoublyValid, zhang2025enhancedmarginalsensitivitymodel, Yadlowsky2022Bounds, Jin2025FSensitivity} that acknowledge the need for distributional consistency between the true and observed distributions. This consistency is generally enforced by either normalizing on a per-covariate basis or analytically constructing a worst-case consistent distribution that lies within the given uncertainty set, where the worst case is in relation to the bound being derived (usually average treatment effect). Our method instead enforces distributional consistency via a Wasserstein bound on the unknown distribution. At the population level, this bound is equivalent to bounding on a per-covariate basis, \revision{but the bound can also be relaxed to account for settings where the distributions being compared do not have the same support, which can arise in sample approximation settings.}
Additionally, our consistency bound is independent of the objective being optimized, whereas the constructions above are tied to the specific causal estimand whose bound is being derived.

\textbf{Robust prediction under distributional shift.} \citet{shimodaira2000loglikelihood} shows that under covariate shift, reweighting examples by the ratio between test and training densities and maximizing the resulting weighted log-likelihood yields asymptotically optimal estimates of a model's regression coefficients, akin to inverse propensity weighting. In our setting, unobserved confounding means that we cannot exactly know these ratios, motivating worst-case optimization over a causal uncertainty set rather than point-estimate reweighting. \citet{shafieezadeh2015robustregression} show that Wasserstein-robust logistic regression admits an exact reformulation as a norm-regularized logistic regression; \citet{MohajerinEsfahani2018Wasserstein} later generalize this result, establishing that minimax optimization over worst-case distributions with Wasserstein uncertainty sets can be reformulated as tractable convex programs across a broader class of losses. While we leverage Wasserstein distance to establish a consistency constraint, our uncertainty set is defined by propensity score bounds from the MSM rather than direct bounds on the distribution of the observed data, yielding a causally interpretable robustness parameter $\Gamma$ that practitioners can reason about directly. \citet{sahoo2026learningbiasedsample} similarly optimize a predictive model against a worst-case distribution within a $\Gamma$-bounded uncertainty set with odds-ratio form identical to the MSM, but their $\Gamma$ bounds outcome-dependent selection into the observed sample rather than outcome-dependent treatment assignment. \citet{chen2016robustcovariateshiftregression} and  \citet{liu2017robustcovariateshiftprediction} extend minimax prediction to general losses, further validating the robustness-over-reweighting philosophy we adopt.

\textbf{Robust policy learning.} \citet{kallus2021minimax} use MSM-derived uncertainty sets for confounding-robust policy improvement in offline settings, optimizing worst-case policy value. Their target — expected reward under a treatment rule — differs fundamentally from ours. Risk scores are upstream of policy: they produce vulnerability estimates that inform human decision-making rather than prescribe treatment directly, a distinction that matters in social services settings where interpretability and human oversight are desired. Additionally, our uncertainty set includes additional consistency constraints. \revision{We show through an example that using the uncertainty set of \citeauthor{kallus2021minimax} causes the uncertainty set to include distributions incompatible with the observed data, and show experimentally that this leads to worse predictions.}

\subsection{Notation} All random variables are defined as measurable functions on an abstract probability space $(\Omega, \mathcal F, {\mathbb P})$ and are capitalized (e.g., $X$), while their realizations are denoted by the same symbols in lowercase letters (e.g.,~$x$). For any random vector $X$, $\mathbb P_{\operatorname{x}}$ denotes its marginal distribution. All (in)equalities involving random variables should be understood to hold almost surely. The family of all bounded Borel-measurable functions from a Borel set $\mathcal A \in \mathcal B(\mathbb R^\ell)$ to another $\mathcal C \in \mathcal B(\mathbb R^k)$ is denoted by $\mathcal L(\mathcal A, \mathcal C)$. For a finite set $\mathcal Y$, $\Delta(\mathcal Y)$ denotes the probability simplex over $\mathcal Y$. The indicator function is denoted by $\mathds{1}(\cdot)$, taking the value $1$ if the argument is true, $0$ otherwise. Finally, $\|\cdot\|$ refers to the Euclidean norm, with the understanding that other choices of norms do not change the theoretical conclusions of the work. 

\section{Problem Description}

Individuals are characterized by their observed covariates $X \in \mathcal{X} \subseteq \mathbb{R}^{d_{\operatorname{x}}}$, unobserved confounders $U \in \mathcal{U} \subseteq \mathbb{R}^{d_{\operatorname{u}}}$, and potential outcomes $\{Y^t\}_{t \in \mathcal T} \in \mathcal Y = \{-1, 1\}^{m+1}$, where $\mathcal T = \{0, 1, \dots, m\}$ is the set of treatments, and $Y^t$ is the potential outcome under $t \in \mathcal T$ \citep{hernan2023causal}. We consider $Y^t = -1$ to be the \textit{adverse outcome}, and use $t=0$ to denote the no-treatment case. We denote by $\mathbb P$ the distribution of $(X, U, \{Y^t\}_{t \in \mathcal T})$.

The distribution $\mathbb P$ is unknown. Instead, we have access to historical data of the form $\{x_i, t_i, y_i\}_{i=1}^n$, where $t_i$ is the treatment historically assigned to $i$ according to an (unknown) historical policy $\pi_{\operatorname{h}} \in \mathcal L(\mathcal X \times \mathcal U, \Delta(\mathcal T))$, and $y_i = y_i^{t_i}$. Here, $\pi^t_{\operatorname{h}}(x,u)$ denotes the probability of assigning treatment $t$ to an individual with covariates $x$ and unobserved confounders $u$. In the observed data, neither the confounders $u_i$ nor the counterfactual outcomes $\{y_i^t\}_{t \in \mathcal T\setminus \{t_i\}}$ are available. We make the following assumption throughout.

\begin{assumption}\label{assum:causal assumptions}
We assume (i) consistency ($Y = Y^T$), (ii) positivity ($\pi_{\operatorname{h}}^t(X,U) \geq \delta > 0$ for all $t \in \mathcal T$), and (iii) conditional exchangeability given both observed covariates and unobserved confounders ($Y^t \perp T \mid (X,U)$ for all $t \in \mathcal T$).
\end{assumption}
Assumption \ref{assum:causal assumptions} (i) and (ii) are standard in the causal inference literature \citep{hernan2023causal}. Assumption \ref{assum:causal assumptions} (iii) is not restrictive, unlike the standard conditional exchangeability assumption (a.k.a. unconfoundedness), which requires $Y^t \perp T \mid X$ for all $t \in \mathcal T$. Indeed, Assumption \ref{assum:causal assumptions} (iii) holds without loss of generality because $U$ is defined broadly and may represent any collection of missing confounders required to have conditional exchangeability.

\paragraph{Problem formulation.} We aim to learn the parameters $\beta$ of a risk score $s(x)=\beta^\top x$ that, given observable covariates $x$, estimates the probability of an adverse outcome ($Y^0=-1$) under no treatment ($t=0$) using the logistic link
\begin{equation*}
\operatorname{Pr}_{\beta}(Y^0 = -1 \mid X=x) = (1 + \exp(-\beta^\top x))^{-1}.
\end{equation*}
Under this model, higher risk score corresponds to a higher probability of an adverse outcome. \revision{$\beta$ is assumed to lie in some set $\mathcal B \subseteq \mathbb{R}^{d_{\operatorname{x}}}$, where $\mathcal B$ can be chosen to capture additional constraints such as integrality or sparsity.}

If $\mathbb P$ were known, we could try to solve the following problem to find the coefficients 
that best explain the relationship between $X$ and $Y^0$:
\begin{equation}\label{prob: ideal_problem}
\begin{aligned}
\displaystyle\sup_{\beta\in\mathcal B} \;&\mathbb E_{\mathbb P}\left[\log \operatorname{Pr}_{\beta}(Y^0 \mid X)\right]
\end{aligned}
\end{equation}
The objective function of \eqref{prob: ideal_problem} is the expected log-likelihood under $\mathbb P$. Problem~\eqref{prob: ideal_problem} cannot be solved because $\mathbb P$ is unknown. If the data were fully observable, a standard sample average approximation (SAA) approach would replace the unknown distribution with the empirical distribution induced by the sample. However, this approach is also infeasible with observational data because we observe $y_i^0$ only when $t_i = 0$, and therefore cannot estimate the joint distribution of $(X, Y^0)$ or the expectation in \eqref{prob: ideal_problem} from the data. A common approach in this case is to rely on so-called propensity scores \citep{rosenbaum1983propensity}, i.e., $\mu_i = \mathbb{P}(T = t_i \mid X = x_i, U = u_i)$. However, these scores also cannot be estimated from the data when there are unobserved confounders. For this reason, we establish a robust framework to model this uncertainty. To this end, using standard arguments, we next reformulate \eqref{prob: ideal_problem} in terms of $\mathbb{P}(T = 0 \mid X, U)$.\footnote{All proofs can be found in Appendix \ref{app: proofs}.}
\begin{lemma}\label{lemma: consistency of IPW reweighting}
It holds that
\begin{equation}\label{eq:in lemma consistency of IPW reweighting}
    \begin{aligned}
        &\mathbb E_{\mathbb P}[\log \operatorname{Pr}_{\beta}(Y^0 \mid X)] \\
        = &\mathbb E_{\mathbb P}\left[\frac{\log \operatorname{Pr}_{\beta}(Y \mid X)}{\mathbb P(T = 0 \mid X, U)} \,\bigg\vert\, T=0\right] \mathbb P(T=0).
    \end{aligned}
\end{equation}
\end{lemma}

The right-hand side of \eqref{eq:in lemma consistency of IPW reweighting} can be interpreted as an expectation under a distribution obtained by reweighting the untreated population inversely to its probability of receiving no treatment, thereby correcting for treatment-selection bias. The idea is that this reweighting recovers the distribution of $(X, Y^0)$ by reweighting the conditional distribution of $(X, Y) \mid T = 0$. We make this argument more explicit later in this section.

We can estimate the conditional distribution of $(X,Y) \mid T=0$ from the data. However, $U$ is never observed, and its meaning and support $\mathcal U$ are unknown. As a result, $\mathbb P(T=0 \mid X, U)$ is not identifiable from the data. To address this uncertainty, we adopt a robust formulation for learning risk scores that perform well under all plausible $\mathbb P(T=0 \mid X, U)$. Since neither $U$ nor $\mathcal U$ is known, we introduce a random variable $Z: \Omega \rightarrow [\delta,1]$, representing the unknown $\mathbb P(T=0 \mid X, U)$, and model it as belonging to an uncertainty set $\mathcal S_{\operatorname{z}}$ of measurable functions from $\Omega$ to $[\delta,1]$. We will impose additional structure on $\mathcal S_{\operatorname{z}}$ to ensure consistency with what can be estimated from the observable data and our modeling goals. We are therefore interested in the following robust optimization problem (constant $\mathbb P(T=0)$ can be dropped from the objective function).
\begin{equation}\label{prob: True robust problem}
\begin{aligned}
\displaystyle\sup_{\beta\in\mathcal B} \inf_{Z \in \mathcal S_{\operatorname{z}}}&\mathbb E_{\mathbb{P}}\left[\frac{\log \operatorname{Pr}_{\beta}(Y \mid X)}{Z} \,\bigg\vert\, T=0\right]
\end{aligned}
\end{equation}
This optimization problem should be interpreted as maximizing the objective of \eqref{prob: ideal_problem} under an adversarial realization of $Z$ that lies in our uncertainty set, thereby ensuring robustness under all possible realizations of $Z$.

\paragraph{Uncertainty set.} We require the following conditions on $\mathcal S_{\operatorname{z}}$. These conditions are satisfied by the (unknown) true $\mathbb P(T=0 \mid X, U)$, i.e., when $Z = \mathbb P(T=0 \mid X, U)$.
We require \textit{consistent inflation of the untreated population}, i.e.,
\begin{equation}
    {\mathbb{E}_{\mathbb P}\left[1/Z \mid T=0\right]} = 1/\mathbb P(T=0)\label{eq: True consistent propensity scores constraints}\tag{C1}
\end{equation}
and \textit{consistent coverage of the covariate distribution}, i.e.,
\begin{equation}
    \mathds{W}(\mathbb P_{\operatorname{x}}, \mathbb Q^{\operatorname{z}}_{\operatorname{x}}) = 0. \label{eq: True dist matching constraint}\tag{C2}
\end{equation}
Here, $\mathds{W}$ is the 1-Wasserstein distance induced by $\|\cdot \|$\revision{, intuitively, the amount of work needed to transform one distribution into another by moving probability mass} (formally defined in Appendix \ref{app: wass defn}), and $\mathbb Q^{\operatorname{z}}$ is the $1/Z$-reweighted distribution among the untreated population:
\begin{equation*}
\mathbb Q^{\operatorname{z}}(A)
= \frac{\mathbb E_{\mathbb P}[\mathds{1}(A)/Z \mid T=0]}{\mathbb E_{\mathbb P}[1/Z \mid T=0]}
\quad \forall A\in\mathcal F.
\end{equation*}
Recalling that $Z$ represents the unknown $\mathbb P(T=0 \mid X, U)$, $\mathbb Q^{\operatorname{z}}$ is a distribution obtained by reweighting the untreated population inversely to its probability of receiving no treatment. Note that this is precisely the distribution referred to in the discussion following Lemma~\ref{lemma: consistency of IPW reweighting}; the right-hand side of \eqref{eq:in lemma consistency of IPW reweighting} is exactly equal to $\mathbb E_{\mathbb Q^{\operatorname{z}}}[\log \operatorname{Pr}_{\beta}(Y^0 \mid X)]$ when $Z = \mathbb P(T=0 \mid X, U)$. Put differently, the $1/Z$-reweighting aims to make the untreated population (i.e., the conditional distribution of $(X, Y) \mid T=0$) representative of the entire population (i.e., the distribution of $(X, Y^0)$). 

Condition \eqref{eq: True consistent propensity scores constraints} ensures that the reweighted untreated population is inflated to the correct full-population scale. For example, if the untreated group makes up $20\%$ of the population, so that $\mathbb P(T=0)=0.2$, then the average weight in this group must be $1/0.2=5$ to scale the untreated group back to the full population size. Condition \eqref{eq: True dist matching constraint} ensures that the reweighting is performed so as to recover the estimable full-population distributional property. In particular, even with abundant data, only the marginal distribution of $X$ can be estimated. We use Wasserstein distance to formulate this requirement because, in finite samples, the covariate supports of the treated and untreated groups often do not perfectly overlap. Wasserstein distance can compare distributions with different supports, unlike alternatives such as $\phi$-divergences.  

The next proposition shows that when $Z = \mathbb P(T=0 \mid X, U)$, the two conditions are satisfied, ensuring consistency with the (unobservable) ground truth distribution.
\begin{proposition}\label{prop: motivation for consistency requirements}
    If $Z = \mathbb P(T=0 \mid X, U)$, then \eqref{eq: True consistent propensity scores constraints} and \eqref{eq: True dist matching constraint} are satisfied.
\end{proposition}

We further impose the following \textit{odds ratio bound} condition on $\mathcal S_{\operatorname{z}}$ to characterize the plausible effects of unobserved confounding:
\begin{align}
\Gamma^{-1}\leq\frac{{\mathbb{P}}(T=0|X)}{1- {\mathbb{P}}(T=0|X)} \frac{1-Z}{Z}\leq \Gamma \label{eq: True odds ratio constraint} \tag{C3}
\end{align}

Condition~\eqref{eq: True odds ratio constraint} is based on Tan's marginal sensitivity model \citep{tan2006msm} and has been used to capture uncertainty due to unobserved confounding, including in minimax regret policy optimization \citep{kallus2021minimax}. It requires that the true odds of receiving no treatment cannot differ from the observable odds based only on $X$ by more than a multiplicative factor of $\Gamma \geq 1$. The factor $\Gamma$ quantifies the degree of hidden bias induced by unobserved confounding, and can be calibrated by domain experts \citep{hsu2013calibrating}. $\Gamma = 1$ corresponds to the absence of unobserved confounding, while larger values of $\Gamma$ allow for a greater potential effect of unobserved confounding. 


{To emphasize the importance of Condition \eqref{eq: True dist matching constraint}, consider an example where the true population distribution is such that $\mathbb P(X = -1) = \mathbb P(X = 1) = 0.5$, and $\mathbb P(T = 0 \mid X = -1) = \mathbb P(T = 0 \mid X = 1) = 0.5$. Consider a selection of the unknown $Z$ such that $Z \mid (X = -1) = 0.25$ and $Z \mid (X = 1) = 0.75$. 
Such a $Z$ satisfies \eqref{eq: True consistent propensity scores constraints} and \eqref{eq: True odds ratio constraint}. However, under such a $Z$, we have $\mathbb Q_{\operatorname{x}}^{\operatorname{z}}(X = -1) = 0.75$ and $\mathbb Q_{\operatorname{x}}^{\operatorname{z}}(X = 1) = 0.25$, so this reweighting selection results in a covariate distribution significantly different from the true covariate distribution overall, and it fails to satisfy \eqref{eq: True dist matching constraint}. This failure is important because the point of reweighting the untreated group is to recover the overall distribution across all treatment groups, but we cannot recover even the covariate distribution.}

\paragraph{Uncertainty set augmentation.}
The previous conditions are those that must be satisfied by the true underlying distribution. However, domain knowledge can also be leveraged to impose additional conditions on the uncertainty set when applicable. Frequently, experts are aware of what certain confounders may be, even if they are not recorded. For example, from conversations with domain experts in homelessness, unobserved confounders could capture a history of domestic violence that is revealed in conversation with a case worker (and thus influences the treatment decision) but not recorded in the data. To encode this idea, we consider an additional \textit{historical rationality} condition:
\begin{align}
    \mathbb E_{\mathbb P}[Y^0/Z \mid T=0] \leq \mathbb E_{\mathbb P}[Y^0/{\mathbb{P}}(T=0|X) \mid T=0] \label{eq: Historical rationality} \tag{C4}
\end{align}

Condition~\eqref{eq: Historical rationality} states that historical treatment decisions should, on average, not have been worse than what would be explained using only observed covariates. This reflects the above intuition that unobserved confounders can provide additional information about $\mathbb{E}[Y^0]$ and be used to better prioritize the most vulnerable individuals for treatment. These bounds can be adapted to reflect alternative historical decision logics if needed. For example, one may consider inverting the above condition if domain knowledge changes over time, or dropping the condition if there is no knowledge at all about the nature of the unobserved confounders.

\section{Sample-Based Approximation} 
Given that $\mathbb P$, or even its marginals that can be estimated from data, are unknown, we cannot solve Problem~\eqref{prob: True robust problem} directly. However, unlike Problem~\eqref{prob: ideal_problem}, Problem~\eqref{prob: True robust problem} admits a sample-based approximation. 

The following additional notation will be useful throughout this section. Recall that $\mu_i = \mathbb{P}(T = t_i \mid X = x_i, U = u_i)$ and let $\hat{\mu}_i$ represent an estimate of $\mathbb{P}(T=t_i | X = x_i)$, which can be learned from the data.
Let $\hat{w}_i = 1/\hat{\mu}_i$, $w_i = 1 / \mu_i$ and $w = (w_1, w_2, \dots, w_n)$. As $w$ is unknown and cannot be estimated from data, we treat it as an uncertain parameter, analogous to before, belonging to an uncertainty set that we will define shortly. Lastly, let $\mathcal{N} = \{1, \dots, n\}$ be the index set of the data samples, and let $\mathcal{N}_0 =\{i \in \mathcal N : t_i = 0\}$ be the subset of indices with $t_i=0$. Then, the sample-based approximation of~\eqref{prob: True robust problem} is given by

\begin{subequations}\label{eq:robust_wass_sample}
\begin{align}
\sup_{\beta \in \mathcal B}  \inf_{w \in \mathcal{U}(\widehat{\mathbb{P}}_n; \Gamma, \varepsilon)}\sum_{i \in \mathcal{N}_0} w_i \log \operatorname{Pr}_{\beta}(y_i \mid x_i), \tag{\theparentequation}\label{eq:robust_wass_sample_obj}
\end{align}

where $\widehat{\mathbb{P}}_n = \frac{1}{n}\sum_{i \in \mathcal{N}}\delta_{x_i}$ is the empirical distribution, $\delta_{x_i}$ is the Dirac
point mass at observation $x_i$, and 
\begin{align}
&\mathcal{U}(\widehat{\mathbb{P}}_n; \Gamma, \varepsilon)=
\left\{\begin{array}{ll}w \in\mathbb{R}_+^{n}: \\  \mathbb{Q}=\sum_{i\in \mathcal{N}_0}\frac{w_i}{ {\sum_{k\in \mathcal{N}_0}  w_k}}\delta_{x_i} \\ \frac{1}{n}\sum_{i \in \mathcal{N}_0}{w}_i = 1 \\  \mathds{W}(\widehat{\mathbb{P}}_n, \mathbb{Q}) \leq \varepsilon \\  \Gamma^{-1} \leq \frac{1/\hat{w}_i}{1-1/\hat{w}_i}\frac{1-1/w_i}{1/w_i} \leq \Gamma \;\; \forall i \in \mathcal{N}_0 \\   \sum_{i \in \mathcal{N}_0} y_i w_i \leq \sum_{i \in \mathcal{N}_0} y_i \hat{w}_i \end{array}\right\} \notag
\end{align}
\end{subequations}

for some $\Gamma \geq 1$ and $\varepsilon \geq 0$. $\mathcal{U}(\widehat{\mathbb{P}}_n; \Gamma, \varepsilon)$ encapsulates the conditions posed on the population level uncertainty set: the first two constraints capture \eqref{eq: True dist matching constraint}, the third captures \eqref{eq: True consistent propensity scores constraints}, the fourth captures \eqref{eq: True odds ratio constraint}, and the fifth captures \eqref{eq: Historical rationality}. We relax \eqref{eq: True dist matching constraint} to $\mathds{W}\leq \epsilon$ because, as noted above, empirical supports can differ even without confounding.
The inner infimum problem in~\eqref{eq:robust_wass_sample} is guaranteed to be feasible when $\varepsilon \geq \mathds{W}(\widehat{\mathbb{P}}_n, \widehat{\mathbb{Q}})$, where $\widehat{\mathbb{Q}} = \sum_{i\in \mathcal{N}_0}\frac{\hat{w}_i}{ {\sum_{k\in \mathcal{N}_0}  \hat{w}_k}}\delta_{x_i}$, and the weights $\hat{w}$ are self-normalized, for example by using the Hájek IPW estimator \citep{hajek1971comment, hernan2023causal}, and greater than~1.

By recasting the Wasserstein distance as a transportation problem and employing strong duality arguments, we formulate problem \eqref{eq:robust_wass_sample_obj} as an exponential cone program amenable to off-the-shelf solvers in the following proposition.


\begin{proposition} \label{prop: Convex Formulation of Robust MLE}
Problem~\eqref{eq:robust_wass_sample}, given that the inner infimum is feasible and $\hat w_i \geq 1$ for all $i \in \mathcal N_0$, 
admits an equivalent 
reformulation as the following exponential cone program: 
\end{proposition}
\begin{subequations}
\label{eq:robust_wass_sample_final_prop}
\begin{align}
\sup \;\;& 
\sum_{i \in \mathcal N_0}\hat w_i \left(\Gamma^{-1}\underline{\theta}_i - \Gamma \overline{\theta}_i - \kappa y_i \right)  \notag \\
+ &\sum_{i \in \mathcal N_0} \left((1-\Gamma^{-1})\underline{\theta}_i -(1-\Gamma)\overline{\theta}_i\right) \tag{\theparentequation}\\
- &\varepsilon\tau + \frac{1}{n}\sum_{j\in\mathcal{N}}\varphi_j+n\rho
\notag\\
\text{s.t.} \;\;\,&\log \operatorname{Pr}_{\beta}(y_i \mid x_i) - \underline{\theta}_i + \overline{\theta}_i + \kappa y_i + \frac{\psi_i}{n}  \geq \rho  \;\; \forall i \in \mathcal{N}_0 \notag\\
&\varphi_j + \psi_i \leq\| x_i-x_j\|\tau  \;\; \forall i \in \mathcal{N}_0,j\in\mathcal{N} \notag\\
& \beta \in \mathcal B, \; \underline{\theta}, \overline{\theta} \in \mathbb{R}_+^{n}, \; \tau, \kappa \in \mathbb{R}_+ \notag \\ & \rho\in \mathbb{R}, \; \varphi\in \mathbb{R}^{n}, \; \psi\in \mathbb{R}^{n} \notag
\end{align}
\end{subequations}

Proposition \ref{prop: Convex Formulation of Robust MLE} allows us to leverage off-the-shelf solvers to compute robust risk scores for certain choices of $\mathcal B$. For our experiments we consider continuous risk score coefficients, and implement and solve the exponential cone program using \citet{mosek}. The number of decision variables grows linearly in $\max(d_{\operatorname{x}}, n)$, while the number of constraints grows quadratically in $n$. While this is considered tractable from a theoretical point of view, the quadratic number of constraints can practically create computational issues for datasets larger than those used in our experiments. In this case, the computation time can be improved via standard delayed constraint generation techniques.

\section{Experiments}
To demonstrate the effectiveness of our method, we run experiments on several datasets from the UCI machine learning repository \citep{UCIML}. 
We begin with a standard binary classification dataset with labels $y_i \in \{-1, 1\}$; note that such datasets are not observational and do not contain a treatment assignment component $t_i$. To convert this into a semi-synthetic \textit{observational} dataset, we introduce a binary treatment variable $t_i \in \{0,1\}$, and while doing so, systematically introduce unobserved confounding. \revision{We adapt datasets in this way so that we have access to both the ground truth and a confounding sample for each dataset; to the best of our knowledge there is no pre-existing repository of intentionally confounded observational datasets.} By considering the labels $y_i$ in the dataset to be the outcome $y_i^0$ under no treatment as well as an unobserved confounder, we generate a treatment policy $\pi^0(x, y^0)$ of the form
\begin{equation*}
    \begin{aligned}
        \log\left(\frac{\pi^0(x,y^0)}{1-\pi^0(x,y^0)} \right) = f(x) + \gamma y^0,
    \end{aligned}
\end{equation*}
where $f: \mathcal X \rightarrow \mathbb R$ and $\gamma \geq 0$, which indicates that the probability of no treatment to a unit with fixed $x$ and a non-adversarial outcome $y^0 = 1$ is at least as high as that for the case of an adversarial outcome $y^0 = -1$. We take $f$ to be an affine function of the observed covariates, and in the experiments we sample the corresponding coefficients $\lambda_1, \dots, \lambda_{d_{\operatorname{x}}}$ independently and identically from a uniform distribution, i.e., $\lambda_1, \dots, \lambda_{d_{\operatorname{x}}} \sim \operatorname{U}(-0.1, 0.1)$.

Note that under this policy, if two individuals $i,j$ have identical observable features $x_i=x_j=x$, we have 
\begin{equation}\label{eq:sensitivity model based log of odds ratio formulation}
    \begin{aligned}
        \frac{\pi^0(x,y_i^0)}{1-\pi^0(x,y_i^0)} \frac{1-\pi^0(x,y_j^0)}{\pi^0(x,y_j^0)} = e^{\gamma (y_i^0-y_j^0)}.
    \end{aligned}
\end{equation}
Thus, for two units with the same observable features, the difference in treatment odds relies only on $\gamma$ and the difference in outcomes. Generating treatments in this manner gives us a reasonable intuition of how $\gamma$ relates to $\Gamma$: since the difference in outcomes is bounded above and below by $2$ and $-2$ respectively, we can approximately bound the ratio of the odds under the true propensity score $\mu_i$ and that under the estimated score $\hat{\mu}_i$ from below and above by $e^{-2\gamma}$ and $e^{2\gamma}$, respectively (see Appendix \ref{app:relating gamma} for details).

We choose 15 classification datasets from the UCI repository for our experiments: details on dataset selection and preprocessing can be found in Appendix \ref{app:datasets}. 1500 training points and 500 test points are sampled from each dataset, and historical treatments are generated for the training set according to the above model for various choices of $\gamma$. We estimate propensity scores using logistic regression, gradient boosting, and a support vector machine, and let $\hat{\mu}$ denote the self-normalized scores from the model with the highest classification accuracy on a held-out validation set of 300 points. Specifically, we normalize such that the inverse weights $\hat{w}$ sum up to $n$ and are all at least 1. Setting $\varepsilon$ to be $1.025$ times the Wasserstein distance between the entire training population and the $1/\hat{\mu}$-reweighted distribution among the untreated population, we then compute our robust risk scores by solving problem \eqref{eq:robust_wass_sample_final_prop} with $\mathcal B = \mathbb{R}^{d_{\operatorname{x}}}$.
We additionally run the following methods for comparison:
\begin{itemize}
    \item Full Information: logistic regression on all data with access to 
    counterfactual outcomes.
    \item Direct Method: logistic regression on $t_i = 0$ data with no propensity weighting.
    \item IPW: logistic regression on $t_i = 0$ data, weighted by $\hat{\mu}$.
    \item Doubly Robust\footnote{Not to be confused with robustness to unobserved confounders.} (DR): a combination of direct and IPW estimation \citep{Robins1994DR}.
    \item Robust Baseline: the exponential cone program reformulation of problem \eqref{eq:robust_wass_sample} without Wasserstein or historical rationality constraints, preserving \eqref{eq: True consistent propensity scores constraints} and \eqref{eq: True odds ratio constraint}.
\end{itemize}
We note that the Robust Baseline invokes an uncertainty set that consists purely of the odds-ratio bound, matching the uncertainty set used by \citet{kallus2021minimax} for policy regret minimization. To the best of our knowledge, this is the state of the art benchmark for our framework, as methods from e.g., \citet{Dorn2025DoublyValid} and \citet{zhang2025enhancedmarginalsensitivitymodel} are analytically tailored for bounding treatment effects. A complete set of ablations is reported in Appendix \ref{app:full}.

For coefficients $\beta$ generated by any of the above methods, performance is evaluated using the following metrics:
\begin{itemize}
    \item Log-likelihood improvement \revision{compared to a method that predicts a risk of $1/2$ for all individuals.}
    \revision{A value of} 1 indicates that $\beta$ yields zero log-loss, and 0 or below indicates that $\beta$ is no better or worse than random guessing:
    \[\text{Improvement} = 1 - \frac{\sum_{i \in \mathcal{N}_{\operatorname{test}}} \log \operatorname{Pr}_{\beta}(y_i \mid x_i)}{\mid \mathcal{N}_{\operatorname{test}}\mid \log 0.5}\]
    
    \item Expected calibration error, reported over 10 bins of equal width. \revision{A value of} 0 indicates perfect calibration, while 1 indicates total miscalibration.
    \item Spearman's rank correlation between the rankings generated by $\beta$ and those generated by the test-data-optimal risk scores, computed via logistic regression with full visibility into the test data, including access to all $y_0$ outcomes. \revision{A value of} 1 indicates that $\beta$ and the test-data-optimal scores generate identical rankings, whereas $-1$ indicates completely opposite rankings. \
\end{itemize}

\begin{table*}[t]
\centering
\footnotesize
\setlength{\tabcolsep}{3pt}
\begin{tabular}{@{}ll ccc ccc ccc ccc@{}}
\toprule
\multirow{2}{*}{\shortstack{$\gamma$ \\ Metric}}
& & \multicolumn{3}{c}{0.0} & \multicolumn{3}{c}{1.0} & \multicolumn{3}{c}{1.5} & \multicolumn{3}{c}{2.0} \\
\cmidrule(lr){3-5} \cmidrule(lr){6-8} \cmidrule(lr){9-11} \cmidrule(lr){12-14}
& $\Gamma$ & Imp.$\uparrow$ & Cal.$\downarrow$ & Corr.$\uparrow$
& Imp.$\uparrow$ & Cal.$\downarrow$ & Corr.$\uparrow$
& Imp.$\uparrow$ & Cal.$\downarrow$ & Corr.$\uparrow$
& Imp.$\uparrow$ & Cal.$\downarrow$ & Corr.$\uparrow$ \\
\midrule
\multicolumn{2}{l}{Full information} & $.329$ & $.050$ & $.878$ & $.329$ & $.050$ & $.878$ & $.329$ & $.050$ & $.878$ & $.329$ & $.050$ & $.878$ \\
\midrule
\multicolumn{2}{l}{IPW} & $.255$ & $.061$ & $.787$ & $.127$ & $.146$ & $.673$ & $-.035$ & $.201$ & $.582$ & $-.312$ & $.253$ & $.489$ \\
\multicolumn{2}{l}{Direct} & $\mathit{.266}$ & $\mathit{.059}$ & $\mathit{.798}$ & $.125$ & $.157$ & $.663$ & $-.054$ & $.221$ & $.555$ & $-.322$ & $.279$ & $.455$ \\
\multicolumn{2}{l}{Doubly Robust} & $.249$ & $.066$ & $.776$ & $.144$ & $.139$ & $.661$ & $.017$ & $.187$ & $.573$ & $-.219$ & $.235$ & $.483$ \\
\midrule
\multirow{4}{*}{Robust Baseline} & $e^{1}$ & $.225$ & $.094$ & $.769$ & $.141$ & $.132$ & $.648$ & $.054$ & $.170$ & $.565$ & $-.124$ & $.215$ & $.483$ \\
 & $e^{2}$ & $.091$ & $.177$ & $.727$ & \cellcolor{green!20}$.084$ & \cellcolor{green!20}$.150$ & \cellcolor{green!20}$.622$ & $.065$ & $.156$ & $.548$ & $-.011$ & $.180$ & $.476$ \\
 & $e^{3}$ & $.030$ & $.207$ & $.714$ & $.032$ & $.174$ & $.610$ & \cellcolor{green!20}$.028$ & \cellcolor{green!20}$.160$ & \cellcolor{green!20}$.530$ & $-.018$ & $.167$ & $.459$ \\
 & $e^{4}$ & $.015$ & $.211$ & $.708$ & $.011$ & $.185$ & $.603$ & $.005$ & $.169$ & $.520$ & \cellcolor{green!20}$-.033$ & \cellcolor{green!20}$\mathit{.163}$ & \cellcolor{green!20}$.446$ \\
\midrule
\multirow{4}{*}{Robust} & $e^{1}$ & $.238$ & $.074$ & $.795$ & $\mathit{.174}$ & $\mathit{.119}$ & $\mathit{.700}$ & $.073$ & $.164$ & $.612$ & $-.109$ & $.211$ & $.504$ \\
 & $e^{2}$ & $.186$ & $.109$ & $.768$ & \cellcolor{green!20}$.160$ & \cellcolor{green!20}$.121$ & \cellcolor{green!20}$.697$ & $\mathit{.107}$ & $\mathit{.145}$ & $\mathit{.617}$ & $.004$ & $.177$ & $\mathit{.511}$ \\
 & $e^{3}$ & $.162$ & $.117$ & $.751$ & $.144$ & $.129$ & $.692$ & \cellcolor{green!20}$.098$ & \cellcolor{green!20}$\mathit{.145}$ & \cellcolor{green!20}$.613$ & $\mathit{.014}$ & $.170$ & $.508$ \\
 & $e^{4}$ & $.156$ & $.115$ & $.742$ & $.139$ & $.131$ & $.690$ & $.092$ & $.146$ & $.608$ & \cellcolor{green!20}$.010$ & \cellcolor{green!20}$.167$ & \cellcolor{green!20}$.503$ \\
\bottomrule
\end{tabular}
\caption{Aggregate results across $\gamma$ values for the following metrics: log likelihood improvement over random guess, expected calibration error, and Spearman's rank correlation with optimal policy. Italicized values indicate the best result for that metric within each $\gamma$ group, excluding Full information: higher values are better for Improvement and Rank Corr., while lower values are better for Cal. Error. Results are reported for each choice of $\Gamma$. In all cases, $\Gamma = e^0$ reduces to the IPW method, and are excluded. Shaded cells mark the a priori choice $\Gamma = e^{2\gamma}$ for the Robust Baseline and Robust methods.}
\label{tab:robust_results}
\end{table*}

Results for 15 independent replications per dataset are collected in Table \ref{tab:robust_results}, aggregated via $10\%$ trimmed mean across all runs. Further statistical analysis is reported in Appendix \ref{app:full}. We first note that the full information benchmark has identical results regardless of $\gamma$, since this method has access to counterfactuals and can ignore the effect of confounders on treatment assignment. As $\gamma$ increases, all other methods decrease in performance, as the treatment assignments (and thus the set of observations the methods have access to) become more confounded. We find that in general, setting $\Gamma$ somewhere between $e^\gamma$ and $e^{2\gamma}$ leads to our robust method having the best performance. \revision{For these choices of $\Gamma$, our robust method outperforms the non-full information benchmarks in all metrics; most notably, we see up to a $28.9\%$ relative reduction in calibration error compared to the DR benchmark. We also see up to an $9.8\%$ relative reduction in calibration error compared to the robust baseline, displaying the benefits of distributional consistency and historical rationality constraints.} 

Additional experiments that adjust the treatment generation method are reported in Appendix \ref{app:expmts}. We find that our method still performs well under a noisy nonlinear relation between $y^0$ and log treatment odds, and that \ref{eq: Historical rationality} can be safely ablated in settings where historical decision makers did not use unobserved confounders to properly prioritize those most at risk.

\section{Acknowledgements}
This material is based upon work supported by the National Science Foundation under Award No. 2346058 titled \textbf{\textit{NRT-AI: Integrating Artificial Intelligence and Operations Research Technologies.}} P. Vayanos is funded in part by the National Science Foundation under CAREER grant 2046230 titled \textbf{\textit{Robust, Interpretable, and Fair Allocation of Scarce Resources in Socially Sensitive Settings}}. Any opinions, findings, and conclusions or recommendations expressed in this material are those of the author(s) and do not necessarily reflect the views or opinions of the National Science Foundation. 

\bibliographystyle{abbrvnat}
\bibliography{bib}

\newpage
\ECSwitch

\ECHead{E-Companion}
\setcounter{secnumdepth}{2}
\section{Additional definitions}
\subsection{1-Wasserstein distance \citep{Kantorovich1958}}
\label{app: wass defn}
The Wasserstein distance $\mathds{W}(\mathbb{Q}_1, \mathbb{Q}_2)$ between two distributions $\mathbb{Q}_1$ and $\mathbb{Q}_2$ on $\Xi \subseteq \mathbb{R}^m$ is defined as
\[
\mathds{W}(\mathbb{Q}_1, \mathbb{Q}_2) = \inf_{\mathbb M \in \mathcal M(\mathbb{Q}_1, \mathbb{Q}_2)} \int_{\Xi^2} \|\xi_1 - \xi_2\| \, \mathbb M(\mathrm{d}\xi_1, \mathrm{d}\xi_2),
\]
where $\mathcal M(\mathbb{Q}_1, \mathbb{Q}_2)$ denotes the set of all joint probability distributions of $\xi_1$ and $\xi_2$ with marginals $\mathbb{Q}_1$ and $\mathbb{Q}_2$, respectively.

\section{Proofs}\label{app: proofs} 
\subsection{Proof of Lemma \ref{lemma: consistency of IPW reweighting}} \label{app: consistency of IPW reweighting}
We have
\begin{equation*}
\begin{aligned}
    &\mathbb E_{\mathbb P}[\log \operatorname{Pr}_{\beta}(Y^0 \mid X)]\\
    =& \mathbb E_{\mathbb P}[\mathbb E_{\mathbb P}[\log \operatorname{Pr}_{\beta}(Y^0 \mid X)\mid X,U]] \\
    =& \mathbb E_{\mathbb P}[\mathbb E_{\mathbb P}[\log \operatorname{Pr}_{\beta}(Y^0 \mid X)\mid X,U,T=0]]\\
    =& \mathbb E_{\mathbb P}[\mathbb E_{\mathbb P}[\log \operatorname{Pr}_{\beta}(Y \mid X)\mid X,U,T=0]],
    \end{aligned}
\end{equation*}
where the second and third equalities follow from Assumption \ref{assum:causal assumptions} (iii) and (i), respectively.
We further have
\begin{equation*}\begin{aligned}
    &\mathbb E_{\mathbb P}[\mathbb E_{\mathbb P}[\log \operatorname{Pr}_{\beta}(Y \mid X)\mid X,U,T=0]] \\ 
    =& \mathbb{E}_{\mathbb P}\left[ \frac{\mathbb{E}_{\mathbb P}\left[ \log \operatorname{Pr}_{\beta}(Y \mid X)\mathds{1}(T=0) \mid X, U \right]}{\mathbb{P}(T = 0 \mid X, U)} \right]\\
    =& \mathbb{E}_{\mathbb P}\left[ \frac{\log \operatorname{Pr}_{\beta}(Y \mid X)\mathds{1}(T=0)}{\mathbb{P}(T = 0 \mid X, U)} \right]\\
    =& \mathbb{E}_{\mathbb P}\left[ \frac{\log \operatorname{Pr}_{\beta}(Y \mid X)}{\mathbb{P}(T = 0 \mid X, U)} \,\bigg\vert\, T=0 \right] \mathbb P(T=0),
    \end{aligned}
\end{equation*}
where equalities follow from standard properties of conditional expectations.

\subsection{Proof of Proposition \ref{prop: motivation for consistency requirements}} \label{app: motivation for consistency requirements}
    Suppose that $Z = \mathbb P(T=0 \mid X, U)$. By the definition of conditional expectation, we have
    \begin{equation*}
        {\mathbb{E}_{\mathbb P}\left[1/Z \mid T=0\right]} = \frac{\mathbb{E}_{\mathbb P}\left[\mathds{1}(T=0)/Z\right]}{\mathbb P(T=0)}.
    \end{equation*}
    By $Z = \mathbb P(T=0 \mid X, U)$ and the properties of conditional expectation, we further have
    \begin{equation*}
    \begin{aligned}
        &\mathbb{E}_{\mathbb P}\left[\mathds{1}(T=0)/Z\right] \\
        =& \mathbb{E}_{\mathbb P}\left[\mathbb{E}_{\mathbb P}\left[\mathds{1}(T=0)/Z \mid X, U\right]\right]\\
        =& \mathbb{E}_{\mathbb P}\left[\mathbb{E}_{\mathbb P}\left[\mathds{1}(T=0)/\mathbb P(T=0 \mid X, U) \mid X, U\right]\right]\\
        =& \mathbb{E}_{\mathbb P}\left[(1/\mathbb P(T=0 \mid X, U))\mathbb{E}_{\mathbb P}\left[\mathds{1}(T=0) \mid X, U\right]\right]\\
        =& \mathbb{E}_{\mathbb P}\left[(1/\mathbb P(T=0 \mid X, U))\mathbb P(T=0 \mid X, U)\right] = 1.
    \end{aligned}
    \end{equation*}
 This shows that \eqref{eq: True consistent propensity scores constraints} is satisfied.

 For \eqref{eq: True dist matching constraint}, note that, by the definition of $\mathbb Q^{\operatorname{z}}$, for all $A \in \mathcal B(\mathcal X)$,
 \begin{equation*}
     \mathbb Q^{\operatorname{z}}_{\operatorname{x}}(A) = \frac{\mathbb E_{\mathbb P}[\mathds{1}(X \in A)/Z \mid T=0]}{\mathbb E_{\mathbb P}[1/Z \mid T=0]}.
 \end{equation*}
 Following similar steps to before, we obtain
 \begin{equation*}
     \mathbb E_{\mathbb P}[\mathds{1}(X \in A)/Z \mid T=0] = \frac{\mathbb E_{\mathbb P}[\mathds{1}(X \in A) \mathds{1}(T=0)/Z]}{\mathbb P(T=0)}
 \end{equation*}
 and
  \begin{equation*}
  \begin{aligned}
  &\mathbb E_{\mathbb P}[\mathds{1}(X \in A) \mathds{1}(T=0)/Z] \\
  = &\mathbb E_{\mathbb P}[\mathbb E_{\mathbb P}[\mathds{1}(X \in A) \mathds{1}(T=0)/Z \mid X,U]]\\
  = &\mathbb P(X \in A).
\end{aligned}
  \end{equation*}
  Therefore, 
  \begin{equation*}
  \begin{aligned}
  &\mathbb Q^{\operatorname{z}}_{\operatorname{x}}(A) = \frac{1}{\mathbb E_{\mathbb P}[1/Z \mid T=0]}\frac{\mathbb P(X \in A)}{\mathbb P(T=0)}\\ &= \mathbb P(T=0) \frac{\mathbb P(X \in A)}{\mathbb P(T=0)}
  = \mathbb P(X \in A) = \mathbb P_{\operatorname{x}}(A).
  \end{aligned}
  \end{equation*}
  This equality holds for all $A \in \mathcal B(\mathcal X)$ so the two probability distributions $\mathbb Q^{\operatorname{z}}_{\operatorname{x}}$ and $\mathbb P_{\operatorname{x}}$ are equal. Because of this and by the definition of the Wasserstein distance, \eqref{eq: True dist matching constraint} is therefore satisfied.

\subsection{Proof of Proposition \ref{prop: Convex Formulation of Robust MLE}} \label{app: Convex Formulation of Robust MLE}
As $\widehat{\mathbb{P}}_n$ and $\mathbb{Q}$ are both discrete distributions, the Wasserstein distance between them can be computed via the following optimal transport problem:
\begin{subequations}\label{eq:wass_transport}
\begin{align}
&\inf_{{\zeta}} &&\sum_{i\in\mathcal{N}_0} \sum_{j\in\mathcal{N}} \| x_i-x_j\| \zeta_{ij}
\notag \\
&\operatorname{s.t.}
&& \sum_{i\in\mathcal{N}_0}\zeta_{ij} = \frac{1}{n}
& \forall j\in\mathcal{N}  \notag\\
&&& \sum_{j\in\mathcal{N}}\zeta_{ij} = \frac{1}{n} w_i
& \forall i\in\mathcal{N}_0  \notag\\
&&& \frac{1}{n}\sum_{i\in \mathcal{N}_0} w_i = 1 \label{eq: consistent reweighting sample} \notag\\
&&&  w \in\mathbb{R}_+^{n}, \;
  \zeta\in\mathbb{R}_+^{n \times n}, \notag
\end{align}
\end{subequations}
where $\zeta_{ij}$ represents the amount of probability mass transported from $x_i$ to $x_j$ in order to transform $\mathbb{Q}$ to $\widehat{\mathbb{P}}_n$.
Note that the third constraint above requires that any feasible realization of weights reweight the untreated population to the correct full-population scale, thereby requiring a sample-based version of \eqref{eq: True consistent propensity scores constraints} to hold.

We therefore have the following equivalent reformulation of problem \eqref{eq:robust_wass_sample}.
\begin{subequations}\label{eq:robust_wass_sample_linconst}
\begin{align}
\sup_{\beta\in\mathcal B}\; &\inf_{w, \zeta}
 && \sum_{i \in \mathcal{N}_0} w_i \log \operatorname{Pr}_{\beta}(y_i \mid x_i)
\notag \\
&\operatorname{s.t.}
&& \Gamma^{-1}(\hat w_i - 1) \leq w_i - 1 \leq \Gamma (\hat w_i - 1)
&\forall i \in \mathcal{N}_0 \notag\\
&&& \sum_{i\in\mathcal{N}_0} \sum_{j\in\mathcal{N}} \| x_i-x_j\| \zeta_{ij} \leq \varepsilon \notag\\
&&& \sum_{i\in\mathcal{N}_0}\zeta_{ij} = \frac{1}{n}
\quad &\forall j\in\mathcal{N} \notag\\
&&& \sum_{j\in\mathcal{N}}\zeta_{ij} = \frac{1}{n} w_i
\quad &\forall i\in\mathcal{N}_0 \notag\\
&&& \frac{1}{n}\sum_{i\in \mathcal{N}_0} w_i = 1 \notag\\
&&& \sum_{i \in \mathcal{N}_0} y_i w_i \leq \sum_{i \in \mathcal{N}_0} y_i \hat{w}_i \notag\\
&&&  w \in\mathbb{R}_+^{n}, \;
  \zeta\in\mathbb{R}_+^{n \times n} \notag
\end{align}
\end{subequations}

Notably, the inner infimum is now linear in ${w}$ and ${\zeta}$. Applying strong duality to the inner infimum gives a supremum problem which can be merged with the outer supremum, thus resulting in an exponential cone problem. Introducing dual variables $\underline{\theta}_i, \overline{\theta}_i, \tau, \varphi_j, \psi_i, \kappa$ for each constraint, respectively (noting that the first constraint corresponds to both $\theta$ terms), and replacing the inner infimum with its dual supremum yields the problem stated in Proposition \ref{prop: Convex Formulation of Robust MLE}. 

\section{Experimental Details}
\subsection{Compute resources}
All experiments are implemented on a cloud computing platform with a mixture of the following compute nodes:
\begin{itemize}
    \item AMD EPYC 7513 CPU @ 2.60 GHz, 248 GB RAM
    \item AMD EPYC 7542 CPU @ 2.90 GHz, 248 GB RAM
    \item INTEL XEON Silver 4116 CPU @ 2.10 GHz, 185 GB RAM
    \item INTEL XEON Silver 4116 CPU @ 2.10 GHz, 89 GB RAM
\end{itemize}

\subsection{Datasets}\label{app:datasets}
To choose datasets to run experiments on, we select classification datasets from the UCI repository for which a logistic relation is a decent, but not perfect explanation (i.e., cross-validated log loss between $0.35$ and $\log 2$, the log-loss of random guess). If a classification dataset is already extremely well explained by a logistic relation, then baseline methods that do not leverage propensity scores would already suffice to explain the corresponding observational dataset, with no dependence on the historical treatment policy. Since this would not be compatible with a setting where unobserved confounders influenced both the historical treatment policy and the outcomes, datasets with log loss below $0.35$ are not considered. We thus acquire the following datasets:

Abalone, Adult, Connect-$4$, Covertype, Default of Credit Card Clients, EEG Eye State, Electrical Grid Stability Simulated Data, NHANES Age Prediction Subset, In-Vehicle Coupon Recommendation, Letter Recognition, MAGIC Gamma Telescope, Online News Popularity, Pen-Based Recognition of Handwritten Digits, Spambase, Wine Quality.

For each dataset, target classes are relabeled with $\{-1, 1\}$: datasets with more than 2 classes accordingly have certain classes merged to reflect binary outcomes, choosing to merge similar classes on a per-domain basis when possible. Categorical features are binarized, numerical features are scaled to $[0,1]$, then all features are centered around their mean. Doing this prevents the treatment model from having a very large random intercept on datasets with many features.

\subsection{Approximately relating $\Gamma$ to $\gamma$}\label{app:relating gamma}
Under the treatment generation model of our experimental section, we have
\begin{equation*}
    \begin{aligned}
        \frac{\mu^0_i}{1-\mu^0_i} \frac{1-\hat{\mu}^0_i}{\hat{\mu}^0_i} \approx &\frac{\pi^0(x_i,y_i^0)}{1-\pi^0(x_i,y_i^0)} \frac{1-\mathbb{E}\left[\pi^0(x_i,Y^0)\right]}{\mathbb{E}\left[\pi^0(x_i,Y^0)\right]}\\ \leq &\frac{\pi^0(x_i,y_i^0)}{1-\pi^0(x_i,y_i^0)} \mathbb{E}\left[\frac{1-\pi^0(x_i,Y^0)}{\pi^0(x_i,Y^0)}\right]\\
        =& \mathbb{E}\left[e^{\gamma (y_i^0-Y^0)}\right] \leq e^{2\gamma}
    \end{aligned}
\end{equation*}
and
\begin{equation*}
    \begin{aligned}
        \frac{\mu^0_i}{1-\mu^0_i} \frac{1-\hat{\mu}^0_i}{\hat{\mu}^0_i}  = &1\Bigg/ \frac{1-\mu^0_i}{\mu^0_i} \frac{\hat{\mu}^0_i}{1-\hat{\mu}^0_i}\\
        \approx &1\Bigg/\frac{1-\pi^0(x_i,y_i^0)}{\pi^0(x_i,y_i^0)} \frac{\mathbb{E}\left[\pi^0(x_i,Y^0)\right]}{1-\mathbb{E}\left[\pi^0(x_i,Y^0)\right]} \\ 
        \geq &1\Bigg/\frac{1-\pi^0(x_i,y_i^0)}{\pi^0(x_i,y_i^0)} \mathbb{E}\left[\frac{\pi^0(x_i,Y^0)}{1-\pi^0(x_i,Y^0)}\right] \\
        = &\mathbb{E}\left[e^{-\gamma (y_i^0-Y^0)}\right] \geq e^{-2\gamma},
    \end{aligned}
\end{equation*}
where all expectations are conditional on $X=x_i$, and the inequalities follow from Jensen's inequality and $0<\pi^0<1$. Thus, we expect $e^{2\gamma}$ to be the highest reasonable value (and $e^{-2\gamma}$ the lowest) for $\Gamma$ in the uncertainty set $\mathcal{U}$.

\section{Further experimental results}
\subsection{Ablations and statistical reporting}\label{app:full}
We report the full set of results from the experimental setup in the main paper. For ablations, we consider the removal of \ref{eq: True dist matching constraint}, \ref{eq: Historical rationality}, and both simultaneously (referred to as Robust Basline in the main paper). For statistical reporting, we first aggregate each metric across datasets for each random seed, then compute standard deviations over the aggregate metrics. Doing this restricts the statistical analysis to replication noise - if we were to compute standard deviations over all results separately, the values would be dominated by dataset heterogeneity. 

Results are presented in Table \ref{tab:full}. We find that removing either \ref{eq: True dist matching constraint} or \ref{eq: Historical rationality} degrade performance, as the uncertainty sets become larger, leading to more conservative predictions. In particular, we find that the ablations deteriorate more quickly when $\Gamma$ is poorly calibrated: when $\gamma = 1$, setting $\Gamma = e^4$ causes the methods without \ref{eq: True dist matching constraint} to produce a log likelihood nearly as poor as that of random guessing.

\begin{table*}[t]
\centering
\footnotesize
\setlength{\tabcolsep}{3pt}
\begin{tabular}{@{}ll ccc ccc ccc ccc@{}}
\toprule
\multirow{2}{*}{\shortstack{$\gamma$ \\ Metric}}
& & \multicolumn{3}{c}{0.0} & \multicolumn{3}{c}{1.0} & \multicolumn{3}{c}{1.5} & \multicolumn{3}{c}{2.0} \\
\cmidrule(lr){3-5} \cmidrule(lr){6-8} \cmidrule(lr){9-11} \cmidrule(lr){12-14}
& $\Gamma$ & Imp.$\uparrow$ & Cal.$\downarrow$ & Corr.$\uparrow$
& Imp.$\uparrow$ & Cal.$\downarrow$ & Corr.$\uparrow$
& Imp.$\uparrow$ & Cal.$\downarrow$ & Corr.$\uparrow$
& Imp.$\uparrow$ & Cal.$\downarrow$ & Corr.$\uparrow$ \\
\midrule
\multicolumn{2}{l}{\multirow{2}{*}{Full information}} & $.329$ & $.050$ & $.878$ & $.329$ & $.050$ & $.878$ & $.329$ & $.050$ & $.878$ & $.329$ & $.050$ & $.878$ \\
 &  & \sd{.009} & \sd{.003} & \sd{.011} & \sd{.009} & \sd{.003} & \sd{.011} & \sd{.009} & \sd{.003} & \sd{.011} & \sd{.009} & \sd{.003} & \sd{.011} \\
\midrule
\multicolumn{2}{l}{\multirow{2}{*}{IPW}} & $.255$ & $.061$ & $.787$ & $.127$ & $.146$ & $.673$ & $-.035$ & $.201$ & $.582$ & $-.312$ & $.253$ & $.489$ \\
 &  & \sd{.023} & \sd{.004} & \sd{.021} & \sd{.074} & \sd{.023} & \sd{.044} & \sd{.119} & \sd{.028} & \sd{.052} & \sd{.208} & \sd{.033} & \sd{.055} \\
\multicolumn{2}{l}{\multirow{2}{*}{Direct}} & $\mathit{.266}$ & $\mathit{.059}$ & $\mathit{.798}$ & $.125$ & $.157$ & $.663$ & $-.054$ & $.221$ & $.555$ & $-.322$ & $.279$ & $.455$ \\
 &  & \sd{\mathit{.018}} & \sd{\mathit{.004}} & \sd{\mathit{.019}} & \sd{.065} & \sd{.023} & \sd{.043} & \sd{.112} & \sd{.030} & \sd{.054} & \sd{.183} & \sd{.034} & \sd{.054} \\
\multicolumn{2}{l}{\multirow{2}{*}{Doubly Robust}} & $.249$ & $.066$ & $.776$ & $.144$ & $.139$ & $.661$ & $.017$ & $.187$ & $.573$ & $-.219$ & $.235$ & $.483$ \\
 &  & \sd{.025} & \sd{.006} & \sd{.022} & \sd{.069} & \sd{.018} & \sd{.044} & \sd{.098} & \sd{.023} & \sd{.051} & \sd{.208} & \sd{.029} & \sd{.055} \\
\midrule
\multirow{8}{*}{Robust} & \multirow{2}{*}{$e^{1}$} & $.238$ & $.074$ & $.795$ & $\mathit{.174}$ & $\mathit{.119}$ & $\mathit{.700}$ & $.073$ & $.164$ & $.612$ & $-.109$ & $.211$ & $.504$ \\
 &  & \sd{.019} & \sd{.005} & \sd{.023} & \sd{\mathit{.045}} & \sd{\mathit{.017}} & \sd{\mathit{.040}} & \sd{.079} & \sd{.024} & \sd{.054} & \sd{.146} & \sd{.027} & \sd{.056} \\
 & \multirow{2}{*}{$e^{2}$} & $.186$ & $.109$ & $.768$ & $.160$ & $.121$ & $.697$ & $\mathit{.107}$ & $\mathit{.145}$ & $\mathit{.617}$ & $.004$ & $.177$ & $\mathit{.511}$ \\
 &  & \sd{.022} & \sd{.006} & \sd{.027} & \sd{.034} & \sd{.010} & \sd{.040} & \sd{\mathit{.054}} & \sd{\mathit{.016}} & \sd{\mathit{.057}} & \sd{.096} & \sd{.020} & \sd{\mathit{.053}} \\
 & \multirow{2}{*}{$e^{3}$} & $.162$ & $.117$ & $.751$ & $.144$ & $.129$ & $.692$ & $.098$ & $\mathit{.145}$ & $.613$ & $\mathit{.014}$ & $.170$ & $.508$ \\
 &  & \sd{.017} & \sd{.006} & \sd{.026} & \sd{.028} & \sd{.008} & \sd{.041} & \sd{.047} & \sd{\mathit{.014}} & \sd{.056} & \sd{\mathit{.081}} & \sd{.018} & \sd{.054} \\
 & \multirow{2}{*}{$e^{4}$} & $.156$ & $.115$ & $.742$ & $.139$ & $.131$ & $.690$ & $.092$ & $.146$ & $.608$ & $.010$ & $.167$ & $.503$ \\
 &  & \sd{.015} & \sd{.005} & \sd{.027} & \sd{.027} & \sd{.007} & \sd{.041} & \sd{.046} & \sd{.013} & \sd{.055} & \sd{.080} & \sd{.017} & \sd{.056} \\
\midrule
\multirow{8}{*}{Robust w/o C2} & \multirow{2}{*}{$e^{1}$} & $.231$ & $.079$ & $.789$ & $.172$ & $\mathit{.119}$ & $.697$ & $.073$ & $.163$ & $.611$ & $-.110$ & $.210$ & $.499$ \\
 &  & \sd{.021} & \sd{.006} & \sd{.024} & \sd{.046} & \sd{\mathit{.017}} & \sd{.041} & \sd{.080} & \sd{.024} & \sd{.055} & \sd{.148} & \sd{.028} & \sd{.055} \\
 & \multirow{2}{*}{$e^{2}$} & $.132$ & $.120$ & $.735$ & $.126$ & $.132$ & $.680$ & $.088$ & $\mathit{.145}$ & $.602$ & $.003$ & $.173$ & $.496$ \\
 &  & \sd{.031} & \sd{.011} & \sd{.033} & \sd{.037} & \sd{.010} & \sd{.038} & \sd{.053} & \sd{\mathit{.013}} & \sd{.056} & \sd{.089} & \sd{.018} & \sd{.056} \\
 & \multirow{2}{*}{$e^{3}$} & $.081$ & $.124$ & $.704$ & $.074$ & $.149$ & $.661$ & $.050$ & $.150$ & $.586$ & $-.008$ & $.162$ & $.491$ \\
 &  & \sd{.017} & \sd{.013} & \sd{.032} & \sd{.031} & \sd{.010} & \sd{.038} & \sd{.046} & \sd{.010} & \sd{.058} & \sd{.072} & \sd{.011} & \sd{.062} \\
 & \multirow{2}{*}{$e^{4}$} & $.069$ & $.121$ & $.692$ & $.052$ & $.155$ & $.652$ & $.026$ & $.158$ & $.579$ & $-.025$ & $\mathit{.159}$ & $.482$ \\
 &  & \sd{.011} & \sd{.013} & \sd{.030} & \sd{.024} & \sd{.009} & \sd{.033} & \sd{.042} & \sd{.010} & \sd{.057} & \sd{.067} & \sd{\mathit{.007}} & \sd{.064} \\
\midrule
\multirow{8}{*}{Robust w/o C4} & \multirow{2}{*}{$e^{1}$} & $.233$ & $.088$ & $.773$ & $.145$ & $.130$ & $.656$ & $.055$ & $.170$ & $.570$ & $-.121$ & $.215$ & $.488$ \\
 &  & \sd{.020} & \sd{.008} & \sd{.021} & \sd{.044} & \sd{.014} & \sd{.043} & \sd{.073} & \sd{.019} & \sd{.048} & \sd{.142} & \sd{.025} & \sd{.050} \\
 & \multirow{2}{*}{$e^{2}$} & $.151$ & $.142$ & $.745$ & $.122$ & $.137$ & $.641$ & $.088$ & $.155$ & $.566$ & $-.009$ & $.183$ & $.490$ \\
 &  & \sd{.027} & \sd{.010} & \sd{.024} & \sd{.027} & \sd{.006} & \sd{.039} & \sd{.044} & \sd{.010} & \sd{.047} & \sd{.092} & \sd{.017} & \sd{.048} \\
 & \multirow{2}{*}{$e^{3}$} & $.115$ & $.161$ & $.729$ & $.105$ & $.145$ & $.633$ & $.079$ & $.155$ & $.559$ & $.005$ & $.176$ & $.484$ \\
 &  & \sd{.020} & \sd{.008} & \sd{.024} & \sd{.022} & \sd{.006} & \sd{.040} & \sd{.037} & \sd{.009} & \sd{.047} & \sd{.078} & \sd{.014} & \sd{.048} \\
 & \multirow{2}{*}{$e^{4}$} & $.108$ & $.164$ & $.724$ & $.100$ & $.147$ & $.632$ & $.073$ & $.156$ & $.555$ & $.001$ & $.173$ & $.479$ \\
 &  & \sd{.017} & \sd{.007} & \sd{.021} & \sd{.020} & \sd{.006} & \sd{.040} & \sd{.036} & \sd{.008} & \sd{.047} & \sd{.076} & \sd{.013} & \sd{.049} \\
\midrule
\multirow{8}{*}{Robust w/o C2, C4} & \multirow{2}{*}{$e^{1}$} & $.225$ & $.094$ & $.769$ & $.141$ & $.132$ & $.648$ & $.054$ & $.170$ & $.565$ & $-.124$ & $.215$ & $.483$ \\
 &  & \sd{.022} & \sd{.009} & \sd{.022} & \sd{.043} & \sd{.013} & \sd{.043} & \sd{.072} & \sd{.018} & \sd{.048} & \sd{.143} & \sd{.025} & \sd{.049} \\
 & \multirow{2}{*}{$e^{2}$} & $.091$ & $.177$ & $.727$ & $.084$ & $.150$ & $.622$ & $.065$ & $.156$ & $.548$ & $-.011$ & $.180$ & $.476$ \\
 &  & \sd{.038} & \sd{.016} & \sd{.025} & \sd{.030} & \sd{.007} & \sd{.038} & \sd{.043} & \sd{.008} & \sd{.047} & \sd{.084} & \sd{.014} & \sd{.048} \\
 & \multirow{2}{*}{$e^{3}$} & $.030$ & $.207$ & $.714$ & $.032$ & $.174$ & $.610$ & $.028$ & $.160$ & $.530$ & $-.018$ & $.167$ & $.459$ \\
 &  & \sd{.020} & \sd{.008} & \sd{.022} & \sd{.024} & \sd{.010} & \sd{.041} & \sd{.034} & \sd{.007} & \sd{.054} & \sd{.066} & \sd{.008} & \sd{.052} \\
 & \multirow{2}{*}{$e^{4}$} & $.015$ & $.211$ & $.708$ & $.011$ & $.185$ & $.603$ & $.005$ & $.169$ & $.520$ & $-.033$ & $.163$ & $.446$ \\
 &  & \sd{.013} & \sd{.007} & \sd{.021} & \sd{.017} & \sd{.009} & \sd{.038} & \sd{.030} & \sd{.007} & \sd{.055} & \sd{.061} & \sd{.006} & \sd{.052} \\
\bottomrule
\end{tabular}
\caption{Aggregate results (standard deviations below) across $\gamma$ values for the following metrics: log likelihood improvement over random guess, expected calibration error, and Spearman's rank correlation with optimal policy. Italicized values indicate the best result for that metric within each $\gamma$ group: higher values are better for Improvement and Rank Corr., while lower values are better for Cal. Error. Results are reported for each choice of $\Gamma$ for each ablation. In all cases, $\Gamma = e^0$ reduces to the IPW method, and are excluded.}
\label{tab:full}
\end{table*}

\subsection{Different historical treatment policies}\label{app:expmts}
The experiments in the main paper assume a positive linear relation between the unobserved confounder and the log odds ratio of treatment. In this section, we consider two alternative historical policies. In the first setting, the term $\gamma$ is instead sampled from the normal distribution $\mathcal{N}(\gamma, \sigma^2)$ for a given spread $\sigma$. In the second setting, the term $\gamma$ is instead replaced with $-\gamma$ with probability $p$. For each of these experiments, we fix $\gamma = 1$ and $\Gamma = e$, and inspect the effect of the hyperparameters $\sigma$ and $p$ respectively.

Results are reported in Tables \ref{tab: noise} and \ref{tab:flip}. When tuning $\sigma$, we find that adding more noise to the system causes all predictions (except the full information oracle) to degrade in performance, with the Robust method still performing the best. When tuning $p$, we find that Robust performs the best for low $p$, Robust w/o \ref{eq: Historical rationality} performs the best for high $p$, and the baselines perform the best in between. We propose the following intuition: settings with low $p$ are aligned with the historical rationality assumption, and settings with high $p$ would reflect a ``historical irrationality'' setting. When $p \approx 0.5$, the effect of $y^0$ on the historical treatment decision is effectively random noise, leading all Robust methods to be overly conservative.

\begin{table*}[t]
\centering
\footnotesize
\setlength{\tabcolsep}{3pt}
\begin{tabular}{@{}l ccc ccc ccc ccc ccc@{}}
\toprule
\multirow{2}{*}{\shortstack{$\sigma$ \\ Metric}}
& \multicolumn{3}{c}{0.2} & \multicolumn{3}{c}{0.4} & \multicolumn{3}{c}{0.6} & \multicolumn{3}{c}{0.8} & \multicolumn{3}{c}{1.0} \\
\cmidrule(lr){2-4} \cmidrule(lr){5-7} \cmidrule(lr){8-10} \cmidrule(lr){11-13} \cmidrule(lr){14-16}
& I$\uparrow$ & C$\downarrow$ & R$\uparrow$
& I$\uparrow$ & C$\downarrow$ & R$\uparrow$
& I$\uparrow$ & C$\downarrow$ & R$\uparrow$
& I$\uparrow$ & C$\downarrow$ & R$\uparrow$
& I$\uparrow$ & C$\downarrow$ & R$\uparrow$ \\
\midrule
\multirow{2}{*}{Full information} & $.329$ & $.050$ & $.878$ & $.329$ & $.050$ & $.878$ & $.329$ & $.050$ & $.878$ & $.329$ & $.050$ & $.878$ & $.329$ & $.050$ & $.878$ \\
 & \sd{.009} & \sd{.003} & \sd{.011} & \sd{.009} & \sd{.003} & \sd{.011} & \sd{.009} & \sd{.003} & \sd{.011} & \sd{.009} & \sd{.003} & \sd{.011} & \sd{.009} & \sd{.003} & \sd{.011} \\
\midrule
\multirow{2}{*}{IPW} & $.131$ & $.141$ & $.674$ & $.121$ & $.135$ & $.673$ & $.101$ & $.134$ & $.674$ & $.066$ & $.137$ & $.668$ & $.028$ & $.142$ & $.659$ \\
 & \sd{.067} & \sd{.022} & \sd{.046} & \sd{.082} & \sd{.023} & \sd{.043} & \sd{.093} & \sd{.025} & \sd{.040} & \sd{.115} & \sd{.030} & \sd{.047} & \sd{.151} & \sd{.034} & \sd{.055} \\
\multirow{2}{*}{Direct} & $.132$ & $.149$ & $.667$ & $.125$ & $.143$ & $.667$ & $.104$ & $.143$ & $.666$ & $.068$ & $.147$ & $.659$ & $.026$ & $.155$ & $.649$ \\
 & \sd{.060} & \sd{.022} & \sd{.045} & \sd{.068} & \sd{.023} & \sd{.043} & \sd{.079} & \sd{.026} & \sd{.041} & \sd{.103} & \sd{.032} & \sd{.047} & \sd{.137} & \sd{.038} & \sd{.058} \\
\multirow{2}{*}{Doubly Robust} & $.147$ & $.135$ & $.663$ & $.147$ & $.128$ & $.667$ & $.130$ & $.128$ & $.663$ & $.102$ & $.131$ & $.659$ & $.072$ & $.135$ & $.651$ \\
 & \sd{.064} & \sd{.018} & \sd{.047} & \sd{.072} & \sd{.020} & \sd{.041} & \sd{.084} & \sd{.023} & \sd{.039} & \sd{.112} & \sd{.026} & \sd{.047} & \sd{.150} & \sd{.030} & \sd{.055} \\
\midrule
\multirow{2}{*}{Robust} & $\mathit{.175}$ & $\mathit{.116}$ & $\mathit{.698}$ & $\mathit{.169}$ & $\mathit{.114}$ & $\mathit{.698}$ & $\mathit{.160}$ & $\mathit{.115}$ & $\mathit{.691}$ & $\mathit{.133}$ & $\mathit{.123}$ & $\mathit{.678}$ & $\mathit{.105}$ & $\mathit{.129}$ & $\mathit{.665}$ \\
 & \sd{\mathit{.040}} & \sd{\mathit{.016}} & \sd{\mathit{.044}} & \sd{\mathit{.052}} & \sd{\mathit{.018}} & \sd{\mathit{.041}} & \sd{\mathit{.059}} & \sd{\mathit{.020}} & \sd{\mathit{.040}} & \sd{\mathit{.081}} & \sd{\mathit{.026}} & \sd{\mathit{.050}} & \sd{\mathit{.121}} & \sd{\mathit{.031}} & \sd{\mathit{.057}} \\
\multirow{2}{*}{Robust w/o C2} & $.173$ & $\mathit{.116}$ & $.696$ & $.166$ & $.115$ & $.694$ & $.158$ & $.117$ & $.688$ & $.131$ & $.124$ & $.673$ & $.103$ & $.131$ & $.661$ \\
 & \sd{.040} & \sd{\mathit{.016}} & \sd{.044} & \sd{.052} & \sd{.018} & \sd{.044} & \sd{.059} & \sd{.020} & \sd{.041} & \sd{.080} & \sd{.026} & \sd{.051} & \sd{.121} & \sd{.030} & \sd{.060} \\
\multirow{2}{*}{Robust w/o C4} & $.150$ & $.129$ & $.658$ & $.149$ & $.127$ & $.663$ & $.144$ & $.128$ & $.662$ & $.124$ & $.131$ & $.657$ & $.102$ & $.134$ & $.652$ \\
 & \sd{.039} & \sd{.013} & \sd{.044} & \sd{.051} & \sd{.015} & \sd{.042} & \sd{.057} & \sd{.017} & \sd{.041} & \sd{.076} & \sd{.021} & \sd{.047} & \sd{.115} & \sd{.025} & \sd{.054} \\
\multirow{2}{*}{Robust w/o C2, C4} & $.145$ & $.131$ & $.651$ & $.145$ & $.129$ & $.656$ & $.140$ & $.130$ & $.657$ & $.121$ & $.133$ & $.652$ & $.099$ & $.137$ & $.646$ \\
 & \sd{.039} & \sd{.013} & \sd{.046} & \sd{.051} & \sd{.015} & \sd{.041} & \sd{.057} & \sd{.017} & \sd{.040} & \sd{.074} & \sd{.020} & \sd{.047} & \sd{.115} & \sd{.024} & \sd{.054} \\
\bottomrule
\end{tabular}
\caption{Aggregate results (standard deviations below) for different noise parameters $\sigma$, reported at $\gamma = 1$ and $\Gamma = e^1$. When $\sigma = 0$, the results match those from the appropriate rows in Table \ref{tab:full}.}
\label{tab: noise}
\end{table*}

\begin{table*}[t]
\centering
\footnotesize
\setlength{\tabcolsep}{3pt}
\begin{tabular}{@{}l ccc ccc ccc ccc ccc@{}}
\toprule
\multirow{2}{*}{\shortstack{$p$ \\ Metric}}
& \multicolumn{3}{c}{0.2} & \multicolumn{3}{c}{0.4} & \multicolumn{3}{c}{0.6} & \multicolumn{3}{c}{0.8} & \multicolumn{3}{c}{1.0} \\
\cmidrule(lr){2-4} \cmidrule(lr){5-7} \cmidrule(lr){8-10} \cmidrule(lr){11-13} \cmidrule(lr){14-16}
& I$\uparrow$ & C$\downarrow$ & R$\uparrow$
& I$\uparrow$ & C$\downarrow$ & R$\uparrow$
& I$\uparrow$ & C$\downarrow$ & R$\uparrow$
& I$\uparrow$ & C$\downarrow$ & R$\uparrow$
& I$\uparrow$ & C$\downarrow$ & R$\uparrow$ \\
\midrule
\multirow{2}{*}{Full information} & $.329$ & $.050$ & $.878$ & $.329$ & $.050$ & $.878$ & $.329$ & $.050$ & $.878$ & $.329$ & $.050$ & $.878$ & $.329$ & $.050$ & $.878$ \\
 & \sd{.009} & \sd{.003} & \sd{.011} & \sd{.009} & \sd{.003} & \sd{.011} & \sd{.009} & \sd{.003} & \sd{.011} & \sd{.009} & \sd{.003} & \sd{.011} & \sd{.009} & \sd{.003} & \sd{.011} \\
\midrule
\multirow{2}{*}{IPW} & $.215$ & $.098$ & $.744$ & $.255$ & $.066$ & $.777$ & $.250$ & $.065$ & $.795$ & $.215$ & $.095$ & $.768$ & $.133$ & $.137$ & $\mathit{.712}$ \\
 & \sd{.036} & \sd{.012} & \sd{.028} & \sd{.023} & \sd{.005} & \sd{.022} & \sd{.020} & \sd{.005} & \sd{.018} & \sd{.028} & \sd{.011} & \sd{.022} & \sd{.053} & \sd{.018} & \sd{\mathit{.032}} \\
\multirow{2}{*}{Direct} & $.222$ & $.101$ & $.745$ & $\mathit{.262}$ & $\mathit{.065}$ & $.784$ & $\mathit{.260}$ & $\mathit{.063}$ & $\mathit{.805}$ & $.222$ & $.098$ & $\mathit{.770}$ & $.127$ & $.149$ & $.702$ \\
 & \sd{.030} & \sd{.011} & \sd{.027} & \sd{\mathit{.019}} & \sd{\mathit{.005}} & \sd{.022} & \sd{\mathit{.017}} & \sd{\mathit{.004}} & \sd{\mathit{.018}} & \sd{.025} & \sd{.012} & \sd{\mathit{.024}} & \sd{.053} & \sd{.021} & \sd{.037} \\
\multirow{2}{*}{Doubly Robust} & $.215$ & $.096$ & $.733$ & $.249$ & $.069$ & $.767$ & $.248$ & $\mathit{.063}$ & $.786$ & $.226$ & $\mathit{.080}$ & $.762$ & $.167$ & $.119$ & $.704$ \\
 & \sd{.034} & \sd{.008} & \sd{.028} & \sd{.022} & \sd{.007} & \sd{.022} & \sd{.019} & \sd{\mathit{.004}} & \sd{.020} & \sd{.028} & \sd{\mathit{.007}} & \sd{.021} & \sd{.048} & \sd{.015} & \sd{.031} \\
\midrule
\multirow{2}{*}{Robust} & $\mathit{.226}$ & $\mathit{.085}$ & $\mathit{.754}$ & $.245$ & $.074$ & $\mathit{.795}$ & $.234$ & $.079$ & $.785$ & $.202$ & $.108$ & $.731$ & $.137$ & $.149$ & $.650$ \\
 & \sd{\mathit{.026}} & \sd{\mathit{.006}} & \sd{\mathit{.025}} & \sd{.018} & \sd{.004} & \sd{\mathit{.019}} & \sd{.017} & \sd{.006} & \sd{.024} & \sd{.025} & \sd{.011} & \sd{.030} & \sd{.046} & \sd{.020} & \sd{.040} \\
\multirow{2}{*}{Robust w/o C2} & $.221$ & $.086$ & $.751$ & $.239$ & $.078$ & $.790$ & $.226$ & $.084$ & $.778$ & $.195$ & $.112$ & $.721$ & $.131$ & $.152$ & $.642$ \\
 & \sd{.027} & \sd{.007} & \sd{.025} & \sd{.019} & \sd{.005} & \sd{.021} & \sd{.019} & \sd{.005} & \sd{.025} & \sd{.027} & \sd{.012} & \sd{.032} & \sd{.048} & \sd{.020} & \sd{.043} \\
\multirow{2}{*}{Robust w/o C4} & $.201$ & $.101$ & $.729$ & $.234$ & $.088$ & $.769$ & $.237$ & $.083$ & $.782$ & $\mathit{.227}$ & $.086$ & $.760$ & $\mathit{.192}$ & $\mathit{.111}$ & $.702$ \\
 & \sd{.028} & \sd{.008} & \sd{.029} & \sd{.019} & \sd{.006} & \sd{.022} & \sd{.014} & \sd{.005} & \sd{.019} & \sd{\mathit{.019}} & \sd{.006} & \sd{.021} & \sd{\mathit{.029}} & \sd{\mathit{.012}} & \sd{.030} \\
\multirow{2}{*}{Robust w/o C2, C4} & $.194$ & $.105$ & $.721$ & $.226$ & $.095$ & $.762$ & $.229$ & $.088$ & $.777$ & $.221$ & $.090$ & $.752$ & $.189$ & $.113$ & $.693$ \\
 & \sd{.030} & \sd{.008} & \sd{.030} & \sd{.021} & \sd{.007} & \sd{.021} & \sd{.016} & \sd{.006} & \sd{.020} & \sd{.020} & \sd{.006} & \sd{.022} & \sd{.030} & \sd{.012} & \sd{.031} \\
\bottomrule
\end{tabular}
\caption{Aggregate results (standard deviations below) for different negation chances $p$, reported at $\gamma = 1$ and $\Gamma = e^1$. When $p$ = 0, the results match those from the appropriate rows in Table \ref{tab:full}.}
\label{tab:flip}
\end{table*}

\section{Future work}
Several aspects of our framework naturally point to directions for future work. The choice of $\mathcal B$ is intentionally modular, such that additional integrality or sparsity constraints can be included if desired. Adding these additional constraints may require additional computational strategies to efficiently solve the resulting problem. 
We also note that our framework allows for different objective functions beyond maximizing log-likelihood, which points to the possibility of tractably and robustly learning quantities beyond risk scores.

\end{document}